\documentclass{article}

\title{The Limitations of Standardized Science Tests as Benchmarks 
for Artificial Intelligence Research: Position Paper}

\author{
Ernest Davis \\
Dept. of Computer Science \\ New York University \\ New York, NY 10012 \\
{\small davise@cs.nyu.edu}}

\begin{document}

\maketitle

\begin{abstract}
In this position paper, I argue that standardized tests for
elementary science such as SAT or Regents tests are not very good benchmarks
for measuring the progress of artificial intelligence systems in
understanding basic science. The primary problem is that these tests are 
designed to test aspects of knowledge and ability that are challenging
for people; the aspects that are challenging for AI systems are very
different. In particular, standardized tests do not test knowledge that is
obvious for people; none of this knowledge can be assumed in AI systems.
Individual standardized tests also have specific features that are
not necessarily appropriate for an AI benchmark. I analyze the
Physics subject SAT in some detail and the New York State Regents
Science test more briefly.
I also argue that the apparent advantages offered
by using standardized tests are mostly either minor or illusory. The one
major real advantage is that the significance is easily explained to the
public; but I argue that even this is a somewhat mixed blessing.
I conclude by arguing that, first, more appropriate collections of
exam style problems could be assembled, and second, that there are
better kinds of benchmarks than exam-style problems. In an appendix
I present a collection of sample exam-style problems that test kinds of
knowledge missing from the standardized tests.
\end{abstract} 

It has often been proposed that a standardized tests
constitute useful goals for AI systems doing automated scientific
reasoning and informative benchmarks for progress. For example 
Brachman et al. (2005) suggest developing a program that can pass the SATs.
Clark, Harrison, and Balasubramanian (2013) propose a project of passing the
New York State Regents Science Test for 4th graders. Strickland
(2013) proposes developing an AI that can pass the entrance exams for the
University of Tokyo. Ohlsson et al. (2013) evaluated the performance 
ConceptNet system (Havasi, Speer, and Alonso 2007)
on a preprocessed form of the Wechesler Preschool and
Primary Scale of Intelligence test. Barker et al. (2004)
describes the construction
of a knowledge-based system that (more or less) scored a 3 (passing) on
two section of the high school chemistry Advanced Placement test.

In this position paper, I want to discuss specifically the
project of developing AI programs to pass standardized science tests, as a 
step toward
developing AI programs with powerful abilities to reason about science; and
I will argue that focusing narrowly on this goal is not the best way of
advancing AI understanding of basic science, and that this benchmark is
not the best way of measuring progress.

The dangers of focusing on too narrow and idiosyncratic a target are 
well illustrated by Watson. IBM set itself the goal of winning at Jeopardy,
and with huge labors, in an extraordinary {\em tour de force,} succeeded.
However, it is not at all clear what contribution this has made to AI
technology; and it seems to have made no contribution whatever to AI theory.
Having succeeded on Jeopardy, IBM is making a huge effort to adapt it to 
other applications; 
what has been published to date on this effort is not particularly impressive.

(Since I have been involved in the Winograd Schema Challenge
(Levesque, Davis, and Morgenstern 2012), let me parenthetically
clarify a related point there. The Winograd Schema Challenge is intended
as a challenge; eventually, if natural language abilities and the inference
mechanisms they require advance sufficiently, some program will pass it.
But there is no point in {\em working\/} on the Winograd Schema Challenge.
Winograd schemas are not a subject matter; they are not even a natural
category. They are a collection of coreference resolution problems,
designed
to be difficult, that conform to a very specific structure in order to
make sure that they are not easily solved by 
statistical techniques over surface
features. The relevant field to work on is natural language
understanding generally or disambiguation specifically.)

Certainly, a standardized science test is a more meaningful and useful
goal for AI than Jeopardy. The questions that make up these tests involve
much more fundamental knowledge and much deeper forms of reasoning
than the trivia about actors, geography, and so on that make up Jeopardy.
So I would certainly expect that an effort to pass a standardized science
test would be very much more fruitful for AI than Watson, except, of course,
in terms of publicity. 
The problems cannot be solved using statistics over surface features. 
Solving them will almost certainly require
major advances in knowledge representation, reasoning, and diagram 
interpretation and significant advances in natural language interpretation.
Indeed Barker et al. (2004) were able to draw some interesting and fruitful
conclusions for knowledge representation and knowledge engineering from their
project aimed at passing the chemistry AP test.
Nonetheless, I will argue that, if we focus our research too narrowly on passing
standardized tests, we will miss critical fundamental issues that will
come back to bite us later; and that if we make too much of these tests
as benchmarks, it will not be helpful to the field in the long run.
I think there are much better ways to formulate goals and benchmarks.

\section{Testing science that any fool knows}
\label{secAnyFool}
Standardized tests were written as tests for human students, and therefore
emphasize aspects of the subject that human students find difficult. 
In humans, successful mastery of these aspects reasonably reliably indicates
mastery of the subject matter. In many case, however, what is difficult
for humans is easy for AI systems, and vice versa; so mastery of these
subjects by an AI system does not at all indicate mastery of the subject
matter.

In particular, standardized tests generally omit aspects of the subject that
``any [human] fool knows'' because these are not worth testing. But of
course, the computer does not necessarily know them. Nor can it be
assumed that progress in encoding formal science will necessarily bring
with it this basic knowledge. One can easily envision an AI program that
knows and can manipulate the equations of general relativity and quantum
electrodynamics, but has no idea how things work in the ordinary world.
The kind of knowledge I have in mind are things like 
\begin{itemize}
\item You can't fit a
watermelon into a sandwich bag. Moreover, you can't make the
watermelon fit by folding it.
However, you could cut off a piece that will fit in the sandwich bag.
\item You can't see anything if your eyes are closed. You can't see anything
if there is no light. You can feel things even if there is no light.
\item If it's warm in your room and cold outside, and you open the window,
the room will get colder.
\end{itemize}
Many more examples are given in appendix A.1.

The importance of having general knowledge that ``any fool knows'' does not
generally diminish as the sophistication of the science involved increases,
except in some cases where the entire subject matter become so recondite
that there is only a remote connection to anything within the scope of
ordinary knowledge.  This is particularly true in reasoning about 
scientific experiments. For example, understanding a chemical experiment
with test tubes and beakers requires basic knowledge about containers
and liquids --- you can pour liquid from the test tube into the beaker, and
so on --- which is nowhere explicit in the chemistry textbook, and may 
not be tested even implicitly on the chemistry SATs. Certainly it is unlikely
that non-standard variants of these --- e.g. if I hold the test tube upside
down over the table, the contents will spill out --- are tested implicitly
on the SATs.

One might argue that this kind of knowledge will needed as
background in order to answer the questions that arise on the standardized
tests; and therefore a project to pass the standardized tests will
necessarily address this kind of knowledge. As far as I can judge from
my examination of the standardized tests, this is not the case; a program
could pass standardized tests with very little of this basic
knowledge. On the other hand, for scientific understanding generally,
this basic knowledge is critical.
In separate unpublished projects  I have analyzed some of
the commonsense reasoning
needed to understand short passages from a biology textbook and from
a description of a high-school chemistry experiment; these basic issues
arise constantly.

\section{Gaps in specific standardized tests}
\label{secSpecificTests}
The gap discussed in section~\ref{secAnyFool}, of questions aimed at very
basic knowledge, is presumably common to all standardized tests except 
possibly tests addressed to cognitive impairment. There is no point in
asking questions that all test takers will get right. Beyond that, though,
each standardized test has its own gaps, from the point of view of an
AI benchmark, reflecting primarily the fact that the test was not
written with the objective of serving as an AI benchmark. Here, of course,
one has to consider each test individually; different tests have 
different strengths and weaknesses.
In general, if an AI researcher wants to use a particular test as a 
benchmark, she should examine it carefully to determine which of the
aims of her research are met by the test and which are not. I have examined
and will discuss below two particular tests: the SAT subject test in
Physics, and the New York State Regents' 4th grade Science test.

First a general caveat: In the world of education, there is a great deal of 
expertise on these standardized tests, which I do not share; and there
is an immense, often rancorous, literature on the strengths, weaknesses, 
biases and so on of these tests,  which I have not read. (Not the least
of my reservations about using these tests as benchmarks is the danger that
the AI community will be seen as taking one side or another in these debates;
or, conversely, that the politics of these debates will distort the analysis
in the AI community.)
However, all of this literature addresses the question of the merits of
these tests in educating humans, not their merits as AI benchmarks. I
do not want to touch the question of educating and testing people
with a ten-foot pole. None
of my analysis below should in any way be viewed as expressing an opinion
on the merits of the tests in the educational system, 
except for the opinion, I hope 
uncontroversial, that it is best to avoid wrongly-posed questions even
in the human setting.

\subsection{The SAT Subject test in Physics}
\label{secSAT}
Since the actual tests are not available (more about this in 
section~\ref{secAdvantages}), I have used as a proxy Sample Test 1
in (Kaplan, 2014 pp.~291-305) on the presumption that the Kaplan people know
what they're doing. I have analyzed the 75 questions on this test along
5 dimensions: Type of problem, mode of mathematical calculation,
physical domain, type of geometry involved, and use of real-world knowledge.
A summary is shown in table~\ref{tabKaplan}. Details are available on
request.

\begin{table}
{\bf Types of problems:} \\
Prediction (34). Calculate/compare numeric features (10). 
Graph reading/manipulation (7).
Identify 
physical law (5). Diagnosis (5). Identify process (4). Planning (2).
Taxonomy (2). Terminology (2). Other (4). \\ \\

{\bf Mode of calculation} \\
Arithmetic (13). Comparative analysis/Sign calculus (8). 
Symbolic algebra (5). Qualitative behavior
of functions presented graphically (5). Identify geometric feature
matching physical law (3). Trivial symbolic algebra (3). 
Add curves presented graphically (2). Other (8).
No calculations involved (28). \\ \\ 

Three of the problems listed above under ``Symbolic algebra'' can be
solved by dimensional analysis and the process of elimination, 
and one can be solved by qualitative reasoning and the process of 
elimination. \\ \\

{\bf Physical domain:} \\
Newtonian mechanics (15). Electromagnetism (10). Wave theory (10).
Kinematics (9). Nuclear physics and elementary particles (7).
Thermodynamics (6). Circuit analysis (6). Optics (5). Gravity. \&
EM (3). Radioactivity (2). Tension (1). Celestial mechanics (1).  \\ \\

{\bf Geometric reasoning}. 28 problems involved geometric reasoning of
various kinds. I did not find a useful way of dividing these into
categories. \\ \\

{\bf Real world knowledge:} 8 problems involved the integration of real-world
knowledge, as discussed in the text.
\caption{Categorization of problems in a sample SAT physics test}
\label{tabKaplan}
\end{table}

In some respects, this meets the objectives of an AI benchmark strikingly
well. The range of types of problems is particularly impressive; certainly
prediction takes the lion's share, but there is a good case 
to be made that that is reasonable. Also
noteworthy is the comparatively slight dependence on exact arithmetic,
and the large role played by qualitative reasoning of one kind and another. 

Still, it is hardly an ideal test for our purposes. The most important
gap is the small number of problems that draw on real world knowledge.
I should say that I am setting the bar on ``real world knowledge'' rather
high here. (Also, of course, the term is not actually apt --- electrons
and so on are certainly part of the real world --- but I can't think of
a better one.) I exclude references to real objects that might as well
be perfectly abstract; for instance, problem 47 refers to ``a car'', but 
it could just as well be ``a mass''. I also exclude uses of the standard 
denizens of physics problems being used in standard ways: pulleys, blocks,
strings, glass as a venue for reflection and refraction, and so on. I likewise 
exclude electronic components such as resistors, capacitors, batteries, and
so on, being used in a standard way.
If there
were a problem in which the mass of a pulley or a resistor
or the thickness of a string
were important, that would count as ``real world'' (there aren't any).

Those excluded, there are in fact eight problems that invoke real world
knowledge, mostly in very minor ways. 
The issue is important enough that we illustrate with four examples; of
the rest, one is very similar to problem 33, and the other three involve
real world knowledge only slightly, at the level of problem 9.

\begin{quote}
{\bf Problem 9:} Which of the physical principles below might be used to
solve the following problem:
A new soft drink bottle is opened, allowing gas to escape into
the atmosphere. As the gas escapes, how does its degree of disorder change?

(A) first law of thermodynamics (conservation of energy) \\
(B) second law of thermodynamics (law of entropy) \\
(C) ideal gas law \\
(D) heat of fusion and heat of vaporization equation \\
(E) heat engine efficiency
\end{quote}
[I have slightly rearranged the statement of the question for clarity.]

The real world scenario is not actually critical here --- one can solve
the problem just by pattern matching on the phrase ``degree of
disorder'', and to some extent one actually does --- but the
reference to the soft drink is certainly
helpful to the human student in grounding what would otherwise be rather 
obscure. It is not clear that this reference
would be of any help to an AI. 

\begin{quote}
{\bf Problem 16:} You are sitting on a seat facing forward on an airplane
with its wings parallel to the ground. The window shades of the
airplane are closed, and the vibration of the plane is negligible. When you
place your class ring on the end of a necklace chain and hold the other
end in front of you, you notice that the chain and ring hang vertically
and point directly to the floor of the airplane. Which of the following
could be true of the airplane?

I. The airplane is at rest. \\
II. The airplane is moving with a constant velocity. \\
III. The airplane is increasing its speed. \\
IV. The airplane is decreasing its speed.

(A) I only \\
(B) III only \\
(C) I or II, but not III or IV. \\
(D) III or IV, but not I or II. \\
(E) IV only
\end{quote}
This problem relies on a number of elements of real world knowledge; the
test taker has to realize that the floor of an airplane is parallel to
the wings and that the ground is perpendicular to the force of gravity.
He also has to understand what it means to put a ring on one end of a chain
and hold the other end. However, none of the 74 other problems draw on
world knowledge in such a rich way, or anywhere close to it.

\begin{quote}


{\bf Problem 27:} [slightly reworded]. If a positively
charged rod is brought near the knobs of a neutral electroscope, which of the 
following statements is true?

(A) The electroscope can be charged negatively without the positively charged
rod touching the knob and using only a grounding wire. \\
(B) The electroscope can be charged positively without the positively charged
rod touching the knob and using only a grounding wire.  \\
(C) The leaves of the electroscope are negatively charged. \\
(D) The knob of the electroscope is positively charged. \\
(E) The electroscope has a net positive charge.
\end{quote}

Answering the question requires some understanding of 
the internal structure of
the electroscope and of the geometry involved; 
it does not suffice to think of the device as a black box.

\begin{quote}


{\bf Problem 33:} [slightly reworded]. A positive charge is moving along
a clockwise circle in the plane of the exam paper you are looking at. The 
direction of the 
magnetic field in the region shown
is  \\
(A) out of the page and perpendicular to it. \\
(B) into the page and perpendicular to it. \\
(C) toward the top of the page. \\
(D) toward the top of the page. \\
(E) to the right.
\end{quote}
Problem 75 is of a similar flavor.

The need for world knowledge here is in interpreting the phrase ``out of
the page''. In itself, this would pose an extraordinarily difficult
problem of interpretation.
(One can imagine the poor AI, mulling over the PDF input, with no
$\hat{z}$ dimension, let alone an outward direction associated with
the content, wondering what on earth the phrase could mean.)
However, since the phrase is common in this kind
of test (and, in fact, apparently associated only with electromagnetic theory
on the SAT exam)
and since there do not seem
to be any other references to the exam page as a physical object, presumably
the interpretation of this in terms of three dimensional geometry would be
hard-coded.\footnote{It would be interesting to know whether human students
have any trouble with this. It is certainly a remarkable example of shifting
the level of abstraction.} 

Thus, the integration of real-world knowledge is rare and often
shallow.\footnote{There may be good reasons for this, in terms of the
human test. Including problems with real-world knowledge can make it
harder to be sure that the physical analysis is correct (we will see an
example in section~\ref{secRegents}); it can introduce unfair biases 
(e.g. does problem 16 give an unfair advantage
to students who have flown in an airplane?) and so on.}
It seems to me that for an AI benchmark one would wish to see a great deal 
more such problems, and problems that require deeper levels of integration.
I give some examples in appendix A.2.

Another category of problem that is underrepresented is problems that require
integrating knowledge from multiple physical domains. In this sample
exam, there is only one example:\footnote{There are two other problems that
require knowledge of both gravity and electromagnetism. However
these do not require {\em integrating} this knowledge; they appear in
separate answers of the multiple choice.}

\begin{quote}
{\bf Question \ldots 38} relate[s] to the two masses $M_{1}$ and $M_{2}$
which have a charge $Q_{1}$ and $Q_{2}$ respectively. The masses are
initially separated by a distance $r$.

If the two charged masses are placed in space so that no other forces affect
them, and they remain at a distance $r$ apart indefinitely, which of
the following must be true? \\
(A) Both charges are positive. \\
(B) $Q_{1}$ is positive and $Q_{2}$ is negative.\\
(C) $Q_{1}$ is negative and $Q_{2}$ is positive. \\
(D) $Q_{1} = Q_{2}$. \\
(E) $M_{1} = M_{2}$.
\end{quote}

This is one of a collection of questions that discuss electric repulsion
and gravitational attraction, so it is clear to the student here that
the point is that these two forces are exactly balanced in the proposed
scenario. The intended answer is thus (A); this is incorrect, since it
is equally possible that both charges are negative.

Again one would wish to see more problems of this kind; 
there are quite a few examples in appendix A.2.

\subsection{Regents' fourth-grade science test}
\label{secRegents}
No one could accuse the Regents' fourth-grade science test of neglecting
real world knowledge; practically all the problems
on the test involve
real world knowledge. 

The Regents science test ranges over a number of subject but gives particular
emphasis to biology. Of the 45 questions on the 2014 exam, 23 were in
biology, 13 in physics, 7 in earth sciences, and 2 in astronomy (all these 
terms being broadly construed). The biology questions do address quite
fundamental issues; the gap in ``what any fool knows'' questions discuss
in section~\ref{secAnyFool} is significantly less acute there, though not
entirely gone. Appendix A.1 here includes a number of biology questions of
various kinds that are
more basic than any on the Regent's test (consider, for example, questions
1.24 through 1.35). The coverage of basic physics, on the
other hand, is not adequate for AI purposes, though it may be suitable
for the fourth grade curriculum. Few physics problems are included,
and those few are rather random (e.g. one problem tests knowledge of 
the fact that a black object gets
hotter in the sun than a white object --- hardly a very fundamental
fact) and somewhat error-prone (see below).

Clark (2015), who advocates the use of this test as a challenge problem for
AI gives three examples of problems that require complex world knowledge.
One of these, indeed, seems to me very much the kind of problem we are 
looking for:

\begin{quote}
A student puts two identical plants in the same type and amount of soil.
She gives them the same amount of water. She puts one of these plants
near a sunny window and the other in a dark room. This experiment tests how
the plants respond to (A) light (B) air (C) water (D) soil. 
\end{quote}
Clark remarks of this that, 
``a reliable answer requires recognizing a model of experimentation
(perform two tasks differing in only one condition), knowing that 
being near a sunny window will explose the plant to light, and that
a dark room has no light in it.'' His description seems correct to me.

However the other two problems that Clark cites seem to be problematic
in various respects. One is this:

\begin{quote}
{\bf Question 13}, 2011 Regents exam

A student riding a bicycle observes that it moves faster on a smooth
road than on a rough road. This happens because the smooth road has \\
A. less gravity \\
B. more gravity \\
C. less friction \\
D. more friction
\end{quote}

The physics here is wrong. A bicycle can hardly be ridden at all
if there is very little friction with the road (on ice, for instance). Once
the friction is sufficient that the bicycle does not slip, increasing
it further should hardly matter, since the bicycle wheel rolls without
sliding along the road. At the contact point, the horizontal velocity
of the wheel is zero.
The fact that it is hard to ride on a rough surface must have
to do with there being too much up and down. Both a student and an AI
system would be penalized for knowing too much physics.

The third problem is similar.

\begin{quote}
Fourth graders are planning a roller-skate race. Which surface would be
the best for this race? (A) gravel (B) sand (C) blacktop (D) grass
\end{quote}

Clark says of this, ``[A] ... reliable answer requires knowing that
a roller-skate race involves roller skating, that roller skating is
on a surface, that skating is best on a smooth surface, and that blacktop
is smooth.'' 

Even aside from the fact that I personally never heard the term ``blacktop''
when I was a child (it was always called ``asphalt'' where I grew up), 
this question and Clark's analysis seem problematic to me. The reason that
sand is unsuited to roller skating on is not so much that sand is not 
smooth --- sandboarding, for example, is much more popular than asphalt
surfing --- but that roller skates tend to sink in sand, and that the sand 
can damage the bearings of the skates. I suspect that children who answer
this question are largely basing it on direct experience of roller skating
or similar activities on different surfaces, rather than using complex
reasoning; and that an AI program would probably do something similar 
(e.g. compare the frequency of school child roller skate races in the various
surfaces that it can find on the web.)

The Regents test also occasionally includes problems
that are seriously badly
formulated. For example, question 32 on the  2014 test read as follows:

\begin{quote}
The data table shows the air temperature at noon for a city in New York
State on five Wednesdays during the month of March. The temperature for 
March 31 has been left blank.

\begin{center}
\begin{tabular}{|l|l|} \hline
Date        & Air Temperature at Noon \\ \hline
March 3     & 42 \\ \hline
March 10    & 45 \\ \hline
March 17    & 48 \\ \hline
March 24    & 51 \\ \hline
March 31    &  ? \\ \hline
\end{tabular}
\end{center}

Based on the pattern shown in the data table, predict the 
air temperature at noon on Wednesday, March 31.
\end{quote}

I must say, I find it appalling that the Regents' exam would include
a question of this kind.
It is simply teaching children that what they learn in school bears no 
relation to the actual world as they know it.\footnote{Science
students of all ages are prone to this misconception. Lambert (2012) tells
of one Harvard physics student who, confronted with David Hestenes' test
of basic physics concepts, asked ``How should I answer these questions 
--- according to what you taught me, or how I usually think about these 
things?''}

The problem with this from the AI perspective is that it is much easier
to write an AI program that gets the intended answer of 54 here than
to write one that realizes that the question is nonsense. All the
program needs is a rule that, if it sees an arithmetic sequence followed
by a blank, it should fill in the next term of the sequence. It does not
have to understand anything else about the question. 
If our 
benchmarks include a lot of problems of this kind, the strong temptation
will be to develop programs that are good test-takers, rather than
good physical reasoners. These kinds of
test-taking tricks have nothing more to do with physical reasoning than
the array of techniques that were programmed into Watson to deal with
Jeopardy puns have to do with natural language understanding.

Finally, this test to a rather surprising extent uses very similar
problems from one year to the next. If a training corpus and a test corpus
are extracted from a corpus collected over several years, the degree of
success is likely to be unrealistically high.

Overall, taking into account both my own and Clark's analysis of these tests,
my feelings about using the Regent's test as a challenge test for AI is 
as follows:
\begin{itemize}
\item Certainly the problems are a challenge to the state of the art in
natural language understanding, and are beyond the state of the art in
diagram interpretation.
\item As a test for an AI system's understanding of basic biology, it has
some value, though many basic aspects of biological knowledge are not
tested either directly or implicitly (e.g. the fact that a cow who has died
will never live again; or that a mammal is born where its mother happens
to be at the time, but not necessarily where its father happens to be.)
\item As a test for an AI system's understanding of basic physics, it is
entirely inadequate. The questions are few; the facts tested are arbitrary
and comparatively unimportant; and the quality control on correctness is poor.
\end{itemize}

\section{Apparent advantages of standardized tests}
\label{secAdvantages}

It seems to me that
the apparent advantages of using standardized tests as
benchmarks instead of
creating our own tests for the purpose and using those
are mostly
either minor or illusory. The advantages that I aware of are the following:

1. Standardized tests exist, in large number; they do not have to
be created. This ``argument from laziness'' is not entirely to be sneezed at.
The experience of the computational linguistics community shows that, if
you take evaluation seriously, developing adequate evaluation
metrics and test materials requires a very substantial effort. On the other
hand, the experience of the computational linguistic community also suggests
if you take evaluation seriously, this effort cannot be avoided by using
preexisting materials. No one in the computational linguistics
community would dream of proposing that NLP programs of any kind should be 
evaluated in terms of their scores on the English language SATs.

2. Entrusting the issue of evaluation measures and benchmarks to the same
physical reasoning community that is developing the programs to be evaluated
is putting the foxes in charge of the chicken coops. The AI researchers will
develop problems that fit their own ideas of how the problems should be 
solved.  This is certainly a legitimate concern;
but I expect in practice much less distortion will be introduced this way than
by taking tests developed for testing people and applying them to AI. Again,
the computational linguistic community has not generally been much troubled
by this issue. (I have heard it argued that the near universal use of the
BLEU measure for machine translation has distorted research in that area.)

3. The standardized tests have been carefully vetted and the performance
of the human population on them is very extensively documented. On the
first point, as we have seen, the vetting does not seem to be completely 
air-tight; a number of questions with bugs have gotten through.
On the second point, there
is no great value to the AI community in knowing how well humans of
different ages, training and so on do on this problem. It hardly matters
which questions can be solved by 5 year olds, which by 12 year olds, and
which by 17 year olds, since, for the foreseeable future, all AI programs
of this kind will be idiot savants (when they are not simply idiots);
capable of superhuman calculations at one minute, and subhuman 
confusions at the next. There is no such thing as the mental age
of an AI program; the abilities and disabilities of an AI program do
not correspond to those of any human being who has ever existed or 
could ever exist.

4. Success on standardized tests is easily accepted by the public (in the
broad sense, meaning everyone except researchers in the area), whereas
success on metrics we have defined ourselves requires explanation, and 
will necessarily be suspect.  This, it seems to me, is the one serious
advantage of using standardized tests. Certainly the public is likely
to take more interest in the 
claim that your program has passed the SAT, or even the fourth-grade Regents
test, than in the claim that it has passed a set of questions that you yourself
designed and whose most conspicuous
feature is that they are spectacularly easy.

However, this is a double-edged
sword. The public can easily jump to the conclusion that, since
an AI program can pass a test, it has the intelligence of a human that
passes the same test.
For example, Ohlsson et al. (2013) entitled their paper
``Verbal IQ of a Four-Year Old Achieved by an AI System.''\footnote{They
have since changed the title to ``Measuring an Artificial Intelligence
System's Performance on a Verbal IQ Test for Young Children''.}
Unfortunately, this was widely misinterpreted in the text as a claim
about verbal intelligence or even general intelligence.
Thus, an
article in {\em ComputerWorld} (Gaudin 2013) had
the headline ``Top Artificial Intelligence System is as smart as a 
4-year-old''; the Independent published an article ``AI System found
to be as clever as a young child after taking IQ test''; and articles
with similar titles were published in many venues.
These headlines are
of course absurd; a four-year old can make up
stories, chat, occasionally follow directions, invent words, learn
language at an incredible pace; ConceptNet (the AI system in question) can
do none of these.

Finally, some standardized tests, including the SAT's, are not published
and are available to researchers only under stringent non-disclosure
agreements. It seems to me that AI researchers should under no circumstances
use such a test with such an agreement. The loss from the inability to
discuss the program's behavior on specific examples far outweighs the
gain from using a test with the imprimatur of the official test designer.
This applies equally to Haroun and Hestenes' (1985) well-known basic physics
test; in any case, it would seem from the published information
that that test focuses on
testing understanding of force and energy rather than testing the
relation of formal physics to basic world
knowledge.

\section{What benchmarks would be better than the standardized tests?}
\label{secBenchmarks}

What benchmarks should be used instead of the standardized tests? If we
want to stick to short answer test formats, which does have advantages,
then certainly the standardized tests are a starting point (once we have
edited out the bad apples.) However, we should supplement those with
many more problems that require real world knowledge; problems that combine
knowledge from different theories;  and problems that use 
forms of reasoning and forms of geometric knowledge that are overlooked
or underutilized.

More importantly, however, we should not confine ourselves to exam-style
benchmarks and goals. Rather, we should consider a variety of tasks such
as:
\begin{itemize}
\item Understanding texts of various kinds: Textbooks, equipment manuals,
text in narrative that draws on physical knowledge (Davis 2013) and so on.
\item Exam questions with essay style answers.
\item Reasoning about variants of physical situations (Davis 1998).
\item Integration with planners in situations that involve complex
physical reasoning.
\item Integration with design programs (Klenk et al. 2014).
\item Guidance for robots.
\end{itemize}

Each of these tasks will require dealing with new aspects of physical
reasoning. The result will be a much richer theory and much more
powerful programs than just looking at answering tests. Of course, designing
benchmarks and metrics for these tasks is harder and less well defined
than for short answer tests. But in general it is better to be working
on the right problem with an imperfect metric than to be working on
the wrong problem with a perfect one.

\section{Final observation}
Standardized tests carry an immense societal
burden and must meet a wide variety of 
very stringent constraints.
They are taken by millions of students annually under very plain
testing circumstances (no use of calculators, let alone Internet).  They
must therefore be gradable either automatically or by not very expert
human graders.  They
bear a disproportionate (and at the current date, ever-increasing) role
in determining the future of those students.  
They must be fair across a wide range of students. They
must conform to existing curricula. They must maintain a constant
level of difficulty, both across the variants offered in any one year,
and from one year to the next. They are subject to intense scrutiny by
large numbers of critics, many of them unfriendly.
These constraints impose serious limitations on what can be asked and 
how exams can be structured.

In developing benchmarks for AI physical reasoning, we are subject to none
of these constraints. Why tie our own hands, by confining ourselves to
standardized tests? Why not take advantage of our freedom?

\section*{References}
K. Barker et al. (2004). ``A Question-Answering System for AP Chemistry:
Assessing KR\&R Technologies,'' {\em KR-2004.}

R.~Brachman et al. (2005). ``Selected Grand Challenges in Cognitive Science,''
MITRE Technical Report 05-1218.

T.L.~Brown, H.E.~LeMay, and B.~Bursten. 2003. {\em Chemistry: The Central
Science,} (9th edn.) Upper Saddle River, NJ: Prentice Hall.

P.~Clark, P.~Harrison, N.~Balasubramanian (2013) ``A Study of the Knowledge
Base Requirements for Passing an Elementary Science Test,'' {\em AKBC-13},

E.~Davis (1998). ``The Naive Physics Perplex,'' {\em AI Magazine,} vol. 19, 
no. 4, Winter 1998, pp. 51-79. 

E.~Davis (2013). ``Qualitative Spatial Reasoning in Interpreting Text and 
Narrative.'' {\em Spatial Cognition and Computation}, {\bf 13}:4, 2013, 264-294.

S.~Gaudin (2013). ``Top Artificial Intelligent system is as smart as
a 4-year old'', {\em Computerworld}, July 15, 2013. 

I.~Haroun and D.~Hestenes (1985). The initial knowledge state of 
college physics students. American Journal of Physics, 53(11), 1043-1055.

C.~Havasi, R.~Speer, J.~Alonso (2007), ``Conceptnet 3: A flexible
multilingual semantic network for common sense knowledge'', 
{\em Recent Advances in Natural Language Processing}, 27-29.

Kaplan (2013). {\em Kaplan SAT Subject Test: Physics. 2013-2014.}
Kaplan Publishing.

M.~Klenk, D.~Bobrow, J.~de Kleer, and B.~Jansenn (2014). 
``Making Modelica Applicable for Formal Methods.''
{\em Proceedings of the 10th International Modelica Conference}.

C.~Lambert (2012). ``Twilight of the Lecture.'' Harvard Magazine, 
March-April, pp. 23-27. 
From http://harvardmagazine.com/2012/03/twilight-of-the-lecture

H.~Levesque, E.~Davis, L.~Morgenstern, (2012). ``The Winograd Schema 
Challenge,'' {\em AAAI-12}.

S.~Ohlsson, R.H.~Sloan, G.~Tur\'{a}n, A.~Urasky (2013), ``Verbal
IQ of a Four-Year Old Achieved by an AI System.'' {\em Commonsense-2013}.

NYSED (2013 and 2014) ``The Grade-4 Elementary-Level Science Test'',\\
http://www.nysedregents.org/Grade4/Science/home.html.

\section*{Appendix: Collection of problems}
Appendix A.1 is a collection of problems, mostly physics, some biology, 
intended to be 
easily solved by fourth-graders. (I am no expert on what realistically
to expect from students of various ages.) Appendix A.2 is a collection
of problems that high school physics students should find easy; they
run from problems whose answer should be immediate to some that
might well require a few minutes' thought. 

The large number of problems about containers reflects the fact that
I have been working for several years about reasoning about containers.
I have included a comparatively large number of problems about astronomy,
first, because astronomy is a very fertile source for problems of qualitative
geometric reasoning and order-of-magnitude reasoning; and second, because 
I like astronomy. I am rather surprised that it is not part of the material
on the physics SATs. There are no questions about electric circuits because
I find them boring; plus they don't lend themselves to commonsense reasoning.

Groups of related problems are bracketed with lines of asterisks.

Please note: The point of this is to give examples of the kinds of
features of problems that I discuss in the main text. It is not intended
in itself as a benchmark collection for AI research, and I do not endorse
its use as such. Still less do I intend it for use with human students.

\subsection*{Appendix A.1: Easy problems}
{\bf Problem 1.1:} You have a bag with some groceries. If you now put a sack of
potatoes into the bag, what will happen? \\
A. The bag will now be lighter. \\
B. The bag will be the same weight. \\
C. The bag will be heavier. 

************************************************************

{\bf Problem 1.2:} You are packing food for a picnic.
You have a big watermelon
which is a foot long and eight inches thick. You have a little 
plastic sandwich 
bag which is four inches wide. Will the watermelon fit in the sandwich bag? 

{\bf Problem 1.3:} Can you make the watermelon fit in the bag by folding the
watermelon? 

{\bf Problem 1.4:} If you have a sharp knife, could you cut a piece of the
watermelon small enough to fit in the bag? 

{\bf Problem 1.5:} If you cut the watermelon into lots of pieces, could you
fit all of them into the bag? 

{\bf Problem 1.6:} Can you put the bag next to the watermelon?

************************************************************

{\bf Problem 1.7:} There is a jar right-side up on a table, with a lid tightly
fastened. There are a few peanuts in the jar.
Joe picks up the jar and shakes it up and down, then puts it back on
the table.
At the end, 
where, probably, are the peanuts? \\
A. In the jar.  \\
B. On the table, outside the jar. \\
C. In the middle of the air. 

{\bf Problem 1.8:} There is a jar right-side up on a table, with a lid tightly
fastened. There are a few peanuts on the table.
Joe picks up the jar and shakes it up and down, then puts it back on
the table.  At the end, where, probably, are the peanuts? \\
A. In the jar.  \\
B. On the table, outside the jar. \\
C. In the middle of the air.

************************************************************

{\bf Problem 1.9:} You are in your room with the door
open. Some music is playing in the next room. You shut the door. Which of
the following is true? \\
A. The sound of music in your room will get softer. \\
B. The sound of music in your room will get louder. \\
C. It won't make any difference.

************************************************************

{\bf Problem 1.10:} You are in your room with the door to the 
rest of the house open
and the window shut.
The whole house is the same temperature, a comfortable $70^{\circ}$F. Outside
it is $40^{\circ}$F. If you open the window, what will happen to the 
temperature in the room? \\
A. The room will get a lot warmer. \\
B. The room will get a lot colder. \\
C. It won't make any difference.

{\bf Problem 1.11:} In the same circumstances as problem 10, what will happen to
the temperature outside? \\
A. It will get a lot warmer outside. \\
B. It will get a lot colder outside. \\
C. It won't make any difference to the temperature outside.

{\bf Problem 1.12:} The situation is the same as problem 10, only
this time, instead of opening the window, you close the door to the rest
of the house. What will happen to the temperature in the room? \\
A. The room will get a lot warmer. \\
B. The room will get a lot colder. \\
C. It won't make any difference to the temperature.

************************************************************

{\bf Problem 1.13:} You and your sister are looking at the moon, 
and then you shut your eyes. Your sister leaves her eyes open.
Can you still see the moon?

{\bf Problem 1.14:}  Can your sister still see the moon?

************************************************************

{\bf Problem 1.15:} Can you see your hand if you hold it in front of your
face?

{\bf Problem 1.16:} Can you see your hand if you hold it behind your head?

{\bf Problem 1.17:} Can you see your hand if it's inside a pocket?

************************************************************

{\bf Problem 1.18:} If you are at home, and you left your jacket at school,
a mile away, can you see your jacket?

{\bf Problem 1.19:} If you are at home, during the day, 
and your jacket is on a chair right
next to you, and you look in its direction, can you see it?

{\bf Problem 1.20:} Suppose you are at home 
and your jacket is on a chair right
next to you and it's nighttime and there's no light at all
in your room. Can you see your jacket, if you look straight at it? 

{\bf Problem 1.21:} Can you feel it, if you touch it? 

{\bf Problem 1.22:} Can you feel it, if you hold your hand close to it
without touching it? 

{\bf Problem 1.23:} If it has some kind of smell, could you smell it, if
you put your nose up to it? 

************************************************************

{\bf Problem 1.24:} After a cat eats a mouse, the mouse is: \\
A. Alive inside the cat's belly. \\
B. Badly injured. \\
C. Dead.

{\bf Problem 1.25:} When a cat eats a mouse, it uses its \\
A. Mouth \\
B. Fur \\
C. Eyes \\
D. Nose

************************************************************

{\bf Problem 1.26:} If a seagull lays eggs that hatch, what will the 
babies be when they grow up? \\
A. Eggs. \\
B. Chickens \\
C. Seagulls. \\
D. Snakes. 

************************************************************

{\bf Problem 1.27:} If a cow dies, how long will it be until the cow is
alive again? \\
A. The cow will be alive again next day.  \\
B. The cow will be alive again in a year. \\
C. The cow will be alive again after her children die. \\
D. The cow will never be alive again.

************************************************************

{\bf Problem 1.28:} If a female eagle and a male alligator have a child,
what would it be? \\
A. Definitely an eagle. \\
B. Definitely an alligator. \\
C. Either an eagle or an alligator. \\
D. A creature that is half an eagle and half an alligator. \\
E. An eagle and an alligator cannot have a child.

************************************************************

{\bf Problem 1.29:} Wolves live in packs. If a wolf gets separated from
its pack, and cannot rejoin the pack then \\
A. It will die within a few hours. \\
B. It will turn into some different kind of animal. \\
C. It will go to sleep until its pack comes back. \\
D. None of the above.

************************************************************

{\bf Problem 1.30:} Sam is a squirrel and Ted and Wendy are his parents.
Ted and Wendy are \\
A. Younger than Sam. \\
B. Exactly the same age as Sam. \\
C. Older than Sam. \\
D. They might be older, or younger, or the same age.

************************************************************

{\bf Problem 1.31:} Fish can only breathe in water. If you are fishing
from a boat and you pull the fish into the boat, then \\
A. It will turn into an animal that can breathe outside of water. \\
B. The boat will fill up with water, so that the fish can breathe. \\
C. The fish will stop breathing but otherwise be OK. \\
D. The fish will die.

************************************************************

{\bf Problem 1.32:} Many birds travel long distances back and forth
every year from their winter home to the summer home and back. If 
you have one of these birds in a zoo, then: \\
A. The bird will stay in the zoo. \\
B. The bird will carry the zoo back and forth from its winter home
to its summer home. \\
C. The bird will escape from the zoo. \\
D. The bird will stay in the zoo. \\
E. The bird will die.

************************************************************

{\bf Problem 1.33:} If a person has a cold, then he will probably get well, \\
A.  In a few minutes. \\
B.  In a few days or a couple of weeks. \\
C.  In a few years. \\
D.  He will never get well.

{\bf Problem 1.34:} If a person cuts off one of his fingers, then he will
probably grow a new finger \\
A.  In a few minutes. \\
B.  In a few days or a couple of weeks. \\
C.  In a few years. \\
D.  He will never grow a new finger.

{\bf Problem 1.35:} If a person hurts himself by stubbing his toe, it should
feel better \\
A.  In a few minutes. \\
B.  In a few days or a couple of weeks. \\
C.  In a few years. \\
D.  It will never feel better.

************************************************************

{\bf Problem 1.36:} Does it hurt to cut your hair?

{\bf Problem 1.37:} Does it hurt if you fall down and scrape your knee?

{\bf Problem 1.38:} Does it hurt if you bang your head against a wall?

{\bf Problem 1.39:} Does it hurt if you lay your head on a pillow?

{\bf Problem 1.40:} Does it hurt if a cat scratches you?

************************************************************

{\bf Problem 1.41:} Suppose that you have two books
a blue book and a red book. The pages are the same size and are made
out of the same kind of paper, but the blue book is much
thicker than the red book. Which, probably, has more pages, the blue book
or the red book? 

{\bf Problem 1.42:} Which is probably heavier, the blue book or the red book? 

{\bf Problem 1.43:} Could you put the red book on top of the blue book? 

{\bf Problem 1.44:} Could you put the red book inside the blue book? 

{\bf Problem 1.45:} Is it possible that there is a page that are
both in the blue book and in the red book? 

{\bf Problem 1.46:} Is it possible that there is a page in the red book
that has exactly the same words as some page in the blue book? 

{\bf Problem 1.47:} Suppose you tear a page out of the blue book, then
tear a page out of the red book, then out of the blue book, then out
of the red book, and so on. What will eventually happen?  \\
A. The blue book will run out of pages, but there will still be pages
in the red book. \\
B. The red book will run out of pages, but there will still be pages
in the blue book. \\
C. Eventually, you will tear the last page out of the blue book, and then
you will tear the last page out of the red book. \\
D. You can keep tearing pages forever.

************************************************************

{\bf Problem 1.48:} Suppose you have two copies of the same book. One has a 
white cover and the other has a black cover, but otherwise they are identical.
Which weighs more? \\
A. The white book weighs more. \\
B. The black book weighs more. \\
C. They weigh the same.

{\bf Problem 1.49:} If you tear a page out of the white book what will happen?
\\
A. The same page will fall out of the black book. \\
B. Another page will grow in the black book. \\
C. The page will grow back in the white book. \\
D. The white book will tear a page out of the black book. \\
E. None of the above.

{\bf Problem 1.50:} If the white book and the black book have a child, what
would it be? \\
A. A black book. \\
B. A white book. \\
C. Either a black or a white book.  \\
D. A book that is half black and half white, \\
E. A grey book. \\
F. Books cannot have children.

************************************************************

{\bf Problem 1.51} Sara has a bucket half full of water. She carefully puts
a couple of stones into the bucket. What happens? \\
A. The stones will float at the top of the water. \\
B. The stones will sink to the bottom of the bucket. \\
C. The stones will sink halfway down. \\
D. The water will all turn into stone. \\
E. The stones will dissolve in the water.

{\bf Problem 1.52} What happens to the level of water in the bucket? \\
A. It gets higher. \\
B. It gets lower. \\
C. It stays the same. \\

************************************************************

{\bf Problem 1.53} George accidentally poured a little bleach into his milk.
Is it OK for him to drink the milk, if he's careful not to swallow
any of the bleach?

************************************************************

{\bf Problem 1.54} When Ed was born, his father was in Boston and his mother
was in Los Angeles. Where was Ed born? \\
A. In Boston. \\
B. In Los Angeles. \\
C. Either in Boston or in Los Angeles. \\
D. Somewhere between Boston and Los Angeles. \\

\subsection*{Appendix A.2: Physics/astronomy problems that should be easy for
high-school physics students}

{\bf Problem 2.1:} You have packed some objects into a 6'' $\times$ 4'' $\times$
8'' box. You have an empty box which is 12'' $\times$ 6 '' $\times$ 12''.
Will the same objects fit into empty box? \\
A. Yes, they will fit. \\
B. No, they will not fit. \\
C. Impossible to tell from the information given . 

{\bf Problem 2.2:} You have packed some objects into a 6'' $\times$ 4'' $\times$
8'' box. You have an empty box which is 6'' $\times$ 6 '' $\times$ 6''.
Will the same objects fit into empty box? \\
A. Yes, they will fit. \\
B. No, they will not fit. \\
C. Impossible to tell from the information given. 

************************************************************

{\bf Problem 2.3:} Suppose that you have a large closed barrel. Empty,
the barrel weighs
1 kg. You put into the barrel 10 gm of water and 1 gm of salt, and you
dissolve the salt in the water. Then you seal the barrel tightly.
Over time, the water evaporates into the 
air in the barrel, leaving the salt at the bottom. If you put the barrel
on a scales after everything has evaporated, the weight will be \\
A. 1000 gm \\
B. 1001 gm \\
C. 1010 gm \\
D. 1011 gm \\
E. Water cannot evaporate inside a closed barrel.

************************************************************

{\bf Problem 2.4:} Does it ever happen that there is an eclipse of the sun
one day and an eclipse of the moon the next?

{\bf Problem 2.5:} Does it ever happen that someone on earth sees an eclipse of
the moon shortly after sunset?

{\bf Problem 2.6:} Does it ever happen that someone on earth sees an eclipse of
the moon at midnight?

{\bf Problem 2.7:} Does it ever happen that someone on earth sees an eclipse of
the moon at noon?

{\bf Problem 2.8:} Does it ever happen that one person on earth sees a total
eclipse of the moon, and at exactly the same time another person sees the 
moon uneclipsed?

{\bf Problem 2.9:} Does it ever happen that one person on earth sees a total
eclipse of the sun, and at exactly the same time another person sees the 
sun uneclipsed?

************************************************************

{\bf Problem 2.10:} Suppose that you are standing on the moon, and the earth
is directly overhead. How soon will the earth set? \\
A. In about a week. \\
B. In about two weeks. \\
C. In about a month. \\
D. The earth never sets. \\

{\bf Problem 2.11:} Suppose that you are standing on the moon, and the sun
is directly overhead. How soon will the sun set? \\
A. In about a week. \\
B. In about two weeks. \\
C. In about a month. \\
D. The sun  never sets. \\

************************************************************

{\bf Problem 2.12:} You are looking in the direction of a 
particular star on a clear night.  
The planet Mars is on a direct line between you and the
star. Can you see the star?

{\bf Problem 2.13:} You are looking in the direction of a particular star on
a clear night.
A small planet orbiting the star is on a direct line between you and
the star. Can you see the star?

************************************************************

{\bf Problem 2.14:} Suppose you were 
standing on one of the moons of Jupiter. Ignoring
the objects in the solar system, which of the following is true: \\
A. The pattern of stars in the sky looks almost identical to the way it
looks on earth. \\
B. The pattern of stars in the sky looks very different from the way
it looks on earth.

************************************************************

{\bf Problem 2.15:} Suppose you are in a room where the temperature is
initially $62^{\circ}$. You turn on a heater, and after half an hour,
the temperature throughout the room is now $75^{\circ}$,
so you turn off the heater. The door to the room is closed; however
there is a gap between the door and the frame, so air can go in and out.
Assume that the temperature and pressure
outside the room remain constant over the time period. Comparing the
air in the room at the start to the air in the room at the end, which of
the following is true:
\begin{quote}
A. The pressure of the air in the room has increased. \\
B. The air in the room at the end occupies a larger volume than the
air in the room at the beginning. \\
C. There is a net flow of air into the room during the half hour period. \\
D. There is a net flow of air out of the room during the half hour period. \\
E. Impossible to tell from the information given.
\end{quote}

{\bf Problem 2.16:} The situation is the same as in problem 65, except that
this time the room is sealed, so that no air can pass in or out. Which of
the following is true:
\begin{quote}
A. The pressure of the air in the room has increased. \\
B. The pressure of the air in the room has decreased. \\
C. The air in the room at the end occupies a larger volume than the
air in the room at the beginning. \\
D. The air in the room at the end occupies a smaller volume than the
air in the room at the beginning. \\
E. The ideal gas constant is larger at the end than at the beginning. \\
F. The ideal gas constant is smaller at the end than at the beginning.
\end{quote}

************************************************************

{\bf Problem 2.17:} You blow up a toy balloon, and tie the end shut. The 
air pressure in the balloon is: \\
A. Lower than the air pressure outside. \\
B. Equal to the air pressure outside.  \\
C. Higher than the air pressure outside.  \\

************************************************************

\vspace*{.2in}


{\bf Problem 2.18:} You have a piston inside a cylinder with a open nozzle
at the other end.
The cylinder is vertical, with the piston at the
bottom and the nozzle at the top. The cylinder is 20 cm high; its
cross section is a circle of radius 3 cm . The radius of the nozzle
is 1/4 cm. You now push upward on the cylinder hard enough so that,
after a fraction of second, 
it moves upward at a constant speed of 5 cm/sec. At the moment when the piston
is 10 cm from the top, how fast is the water moving
when it comes out of the nozzle? \\

{\bf Problem 2.19:} In the situation described in problem 2.18, how high does
the fountain of water go? (Ignore air resistance.) \\

{\bf Problem 2.20:} As the piston approaches the top of the cylinder,
does the speed of the water coming out increase, decrease, or stay the same?\\

{\bf Problem 2.21:} Let $w_{p}$ be the weight of the piston and let
$w_{w}(t)$ be the weight of the water that remains in the piston at time $t$.
Once the piston has reached the speed of 5 cm/sec, let $f(t)$ be the force
on the piston needed at time $t$ to keep it moving at a constant upward
speed. Which of the following is true: \\
A. $f(t)=0$, because the piston is not accelerating. \\
B. $f(t) = w_{p}$. \\
C. $w_{p} < f(t) < w_{p} + w_{w}(t)$. \\
D. $f(t) = w_{p} + w_{w}(t)$. \\
E. $f(t) > w_{p} + w_{w}(t)$. \\

{\bf Problem 2.22:} Suppose you replace the water by a heavier liquid, like 
mercury, but otherwise left the problem the same (the piston still
moves at 5 cm/sec). Which of the following would change: \\
A. The speed of the liquid leaving the piston. \\
B. The height of the fountain of liquid. \\
C. The force needed on the piston.

************************************************************

\vspace*{.2in}



{\bf Problem 2.23:} In the Millikan oil-drop experiment, a tiny oil drop charged
with a single electron was suspended between two charged plates
field. The charge on the plates was adjusted until the electric force
on the drop exactly balanced its weight. How were the plates charged? \\
A. Both plates had a positive charge. \\
B. Both plates had a negative charge. \\
C. The top plate had a positive charge, and the bottom plate had a negative
charge. \\
D. The top plate had a negative charge, and the bottom plate had a positive
charge. \\
E. The experiment would work the same, no matter how the plates were
charged.

{\bf Problem 2.24:} If the oil drop started moving upward, Millikan
would \\
A. Increase the charge on the plates \\
B. Reduce the charge on the plates. \\
C. Increase the charge on the drop.\\
D. Reduce the charge on the drop.\\
E. Make the drop heavier.\\
F. Make the drop lighter.\\
G. Lift the bottom plate.

{\bf Problem 2.25:} If the oil drop fell onto the bottom plate, Millikan would \\
A. Increase the charge on the plates \\
B. Reduce the charge on the plates. \\
C. Increase the charge on the drop.\\
D. Reduce the charge on the drop.\\
E. Start over with a new oil drop.

{\bf Problem 2.26:} The experiment demonstrated that charge is quantized; that
is, the charge on an object is always an integer multiple of the charge
of the electron, not a fractional or other non-integer multiple. To
establish this, Millikan had to measure the charge on \\
A. One oil drop. \\
B. Two oil drops. \\
C. Many oil drops.

************************************************************

Read the following description of a chemistry experiment,\footnote{Do not 
attempt to
carry out this
experiment based on the description here.  Potassium chlorate is explosive,
and safety precautions, not described here, must be taken.} illustrated
below.
A small quantity of potassium chlorate ($\mbox{KClO}_{3}$)
is heated in a test tube, and decomposes into potassium chloride
(KCl) and oxygen ($\mbox{O}_{2}$).
The gaseous
oxygen expands out of the test tube, goes through the tubing, bubbles up
through the water in the beaker, and collects in the inverted beaker over the
the water. Once the bubbling has stopped, the experimenter raises or lowers
the beaker until the level of the top of water inside and outside the beaker
are equal.  At this point, the pressure in the beaker is equal to atmospheric
pressure.
Measuring the volume of the gas collected over the water, and
correcting for the water vapor which is mixed in with the oxygen, the
experimenter can
thus measure the amount of oxygen released in the decomposition.

\vspace*{.2in}


\vspace*{.2in}

{\bf Problem 2.27:} If the right end of the U-shaped tube were outside
the beaker rather than inside, how would that change things? \\
A. The chemical decomposition would not occur. \\
B. The oxygen would remain in the test tube. \\
C. The oxygen would bubble up through the water in the basin to the open
air and would not be collected in the beaker. \\
D. Nothing would change. The oxygen would still collect in the beaker,
as shown.

{\bf Problem 2.28:} If the beaker had a hole in the base (on top when inverted
as shown), how would that change things? \\
A. The oxygen would bubble up through the beaker and out through the hole. \\
B. Nothing would change. The oxygen would still collect in the beaker,
as shown. \\
C. The water would immediately flow out from the inverted beaker into the basin
and the beaker would fill with air coming in through the hole.

{\bf Problem 2.29} If the test tube, the beaker, and the U-tube were all
made of stainless steel rather than glass, how would that change things? \\
A. The chemical decomposition would not occur. \\
B. The oxygen would seep through the stainless steel beaker. \\
C. Physically it would make no difference, but it would be impossible to see
and therefore impossible to measure. 

{\bf Problem 2.30} Suppose the stopper in the test tube were removed, but
that the U-tube has some other support that keeps it in its current position.
How would that change things? \\
A. The oxygen would stay in the test tube. \\
B. All of the oxygen would escape to the outside air. \\
C. Some of the oxygen would escape to the outside air, and some would go
through the U-shaped tube and bubble up to the beaker. So the beaker
would get some oxygen but not all the oxygen.

{\bf Problem 2.31} The experiment description says,
``The experimenter raises or lowers
the beaker until the level of the top of water inside and outside the beaker
are equal.  At this point, the pressure in the beaker is equal to atmospheric
pressure.'' More specifically: Suppose that after the bubbling has stopped,
the level of water in the beaker is higher than the level in the basin
(as seems to be shown in the right hand picture).  Which of the following
is true: \\
A. The pressure in the beaker is lower than atmospheric pressure, and the
beaker should be lowered. \\
B. The pressure in the beaker is lower than atmospheric pressure, and the
beaker should be raised. \\
C. The pressure in the beaker is higher than atmospheric pressure, and the
beaker should be lowered. \\
D. The pressure in the beaker is higher than atmospheric pressure, and the
beaker should be raised. 

{\bf Problem 2.32} Suppose that instead of using a small amount of 
potassium chlorate, as shown, you put in enough to nearly fill the test tube.
How will that change things? \\
A. The chemical decomposition will not occur. \\
B. You will generate more oxygen than the beaker can hold. \\
C. You will generate so little oxygen that it will be difficult to measure.

{\bf Problem 2.33} In addition to the volume of the gas in the beaker, which
of the following are important to measure accurately? \\
A. The initial mass of the potassium chlorate. \\
B. The weight of the beaker. \\
C. The diameter of the beaker. \\
D. The number and size of the bubbles. \\
E. The amount of liquid in the beaker.

{\bf Problem 2.34} The illustration shows a graduated beaker. Suppose instead
you use an ungraduated glass beaker. How will that change things? \\
A. The oxygen will not collect properly in the beaker. \\
B. The experimenter will not know whether to raise or lower the beaker. \\
C. The experimenter will not be able to measure the volume of gas.

{\bf Problem 2.35} The illustration shows two separate basins and beakers.
What is the significance of that? \\
A. These are two separate basins and beakers. \\
B. The left hand picture (a) shows the state of things at the very
start of the experiment; the right hand picture (b) shows the state of
things at the very end of the experiment. \\
C. The left hand picture (a) shows the state of things toward the beginning
of the experiment, after some gas has been evolved and collected;
the right hand picture (b) shows the state of
things toward the end, after the bubbling has stopped but before the
levels have been equalized. \\
D. The right hand picture illustrates safety procedures.

{\bf Problem 2.36} Both pictures (a) and (b) show the mouth of the beaker
below the level of the water in the basin. Suppose that instead the mouth
of the beaker is above the level of water in the basin. What would
happen? \\
A. The water would flow out of the beaker into the basin. \\
B. The water will stay in the beaker, but the 
oxygen would escape into the open air in the gap between the mouth of the
beaker and the surface of water in the basin. \\
C. The oxygen will collect in the beaker, but it will be impossible to carry
out the procedure of balancing the pressures by raising and lowering the
beaker.

{\bf Problem 2.37} At the start of the experiment, the beaker needs to
be full of water, with its mouth in the basin below the surface of the
water in the basin. How is this state achieved? \\
A. Fill the beaker with water right side up, turn it upside down, and
lower it upside down into the basin. \\
B. Put the beaker rightside up into the basin below the surface of the water;
let it fill with water; turn it upside down keeping it underneath the water;
and then lift it upward, so that the base is out of the water, but keeping
the mouth always below the water.\\
C. Put the beaker upside down into the basin below the surface of the water;
and then lift it back upward, so that the base is out of the water, but keeping
the mouth always below the water. \\
D. Put the beaker in the proper position, and then splash water upward
from the basin into it. \\
E. Put the beaker in its proper position, with the mouth below the level
of the water; break a small hole in the base of the beaker; suction the
water up from the basin into the beaker using a pipette; then fix the hole

(What would be really cool here would be to show animations of the five
possibilities.) 

{\bf Problem 2.38} From the time that you first bring the heat to the
test tube to the time that you finish measuring the volume of gas, how
much time would you think elapses? \\
A. A fraction of a second. \\
B. Several minutes to an hour. \\
C. Several days. \\
D. A year or more.

************************************************************

{\bf Problem 2.39} Nearby stars exhibit parallax due to the annual motion
of the earth. If a star is nearby, and is in the plane of the earth's 
revolution, and you track its relative motion against the background of
very distant stars over the course of a year, what
figure does it trace?

{\bf Problem 2.40} If a star is nearby, and the line from the earth to
the star is perpendicular to the plane of the earth's
revolution, and you track its relative motion against the background of
very distant stars over the course of a year, what
figure does it trace?

************************************************************

{\bf Problem 2.41} A star exhibits the following unusual behavior: Every
20 days, it grows gradually dimmer for an hour, stays dim for three hours,
and then over the next hour returns to its usual brightness. 

The following explanation is conjectured: The star has a dark twin which
rotates around it. The dim time corresponds to the time that the dark 
twin is partially occluding the bright star, from the point of view 
of earth.

It is further observed that the cycle time is not quite constant; it is
slightly shorter in the spring and slightly longer in the fall.

Which of the following explanations of this variance in the cycle time is
most plausible?

A. The earth is closer in the spring and further in the fall, so the light
take less time to travel in the spring. \\
B. The earth is moving toward the star in the spring and moving away
in the fall, so in the spring it is closer at each successive observation. \\
C. The earth is closer in the spring and further in the fall. The earth's
gravity is affecting the star's revolution, so that it moves faster
in the spring and slower in the fall. \\
D. There is a third invisible star that is affecting the dark star's behavior.

{\bf Problem 2.42:} This change in cycle time will be largest if \\
A. The star is in the plane of the earth's revolution. \\
B. The star is on the line from the earth perpendicular to the plane
of the earth's revolution. \\
C. The position of the star in the sky makes no difference.

{\bf Problem 2.43:} Suppose that the star is in the plane of the earth's 
revolution and 20 light years away. What is the difference between
the cycle time at its shortest and the cycle time at its longest? Note:
the earth's orbital velocity is about 30 km/sec; the speed of light is
about 300,000 km/sec.

{\bf Problem 2.44:} This effect is analogous to:\\
A. Precesssion of the equinoxes. \\
B. Doppler effect/red shift. \\
C. Motion of a mass on a spring. \\
D. Foucault's pendulum.

\end{document}